\documentclass[conference]{IEEEtran}
\IEEEoverridecommandlockouts
\usepackage{cite}
\usepackage{amsthm}
\usepackage{amsmath,amssymb,amsfonts}
\usepackage{algorithmic}
\usepackage{graphicx}
\usepackage{textcomp}
\usepackage{xcolor}
\usepackage[short]{optidef}

\usepackage{lipsum}

\DeclareRobustCommand*{\IEEEauthorrefmarkNum}[1]{%
  \raisebox{0pt}[0pt][0pt]{\textsuperscript{\footnotesize #1}}%
}

\def\BibTeX{{\rm B\kern-.05em{\sc i\kern-.025em b}\kern-.08em
    T\kern-.1667em\lower.7ex\hbox{E}\kern-.125emX}}
\begin{document}

\title{A Learning-based Incentive Mechanism for Mobile AIGC Service in Decentralized Internet of Vehicles\\
% \thanks{thanks}
}

% \author{\IEEEauthorblockN{Jiani Fan, Minrui Xu, Ziyao Liu, Huanyi Ye, Chaojie Gu, Dusit Niyato, and Kwok-Yan Lam}
% \IEEEauthorblockA{\textit{School of Computer Science and Engineering, Nanyang Technological University, Singapore} \\
% \textit{College of Control Science and Engineering, Zhejiang University, China} \\
% \{jiani001, minrui001, ziyao002, huanyi001\}@e.ntu.edu.sg, gucj@zju.edu.cn, \{dniyato, kwokyan.lam\}@ntu.edu.sg}}

\author{
    \IEEEauthorblockN{%
    Jiani Fan\IEEEauthorrefmarkNum{1},
    Minrui Xu\IEEEauthorrefmarkNum{1}, 
    Ziyao Liu\IEEEauthorrefmarkNum{1},
    Huanyi Ye\IEEEauthorrefmarkNum{1},
    Chaojie Gu\IEEEauthorrefmarkNum{2},
    Dusit Niyato\IEEEauthorrefmarkNum{1},
    Kwok-Yan Lam\IEEEauthorrefmarkNum{1}
    }
    \IEEEauthorblockA{%
    \IEEEauthorrefmarkNum{1}School of Computer Science and Engineering, Nanyang Technological University, Singapore\\
    \IEEEauthorrefmarkNum{2}College of Control Science and Engineering, Zhejiang University, People's Republic of China\\
    \{jiani001, minrui001, ziyao002, huanyi001\}@e.ntu.edu.sg, gucj@zju.edu.cn, \{dniyato, kwokyan.lam\}@ntu.edu.sg
    }
    \thanks{This research is supported by the National Research Foundation, Singapore, and Infocomm Media Development Authority under its Future Communications Research \& Development Programme, DSO National Laboratories under the AI Singapore Programme (AISG Award No: AISG2-RP-2020-019), Energy Research Test-Bed and Industry Partnership Funding Initiative, Energy Grid (EG) 2.0 programme, DesCartes and the Campus for Research Excellence and Technological Enterprise (CREATE) programme, and MOE Tier 1 (RG87/22). Jiani Fan's research is partly supported by Alibaba-NTU JRI Talent Programme, Nanyang Technological University, Singapore.}
}

\maketitle

\begin{abstract}
Artificial Intelligence-Generated Content (AIGC) refers to the paradigm of automated content generation utilizing AI models. Mobile AIGC services in the Internet of Vehicles (IoV) network have numerous advantages over traditional cloud-based AIGC services, including enhanced network efficiency, better reconfigurability, and stronger data security and privacy. Nonetheless, AIGC service provisioning frequently demands significant resources. Consequently, resource-constrained roadside units (RSUs) face challenges in maintaining a heterogeneous pool of AIGC services and addressing all user service requests without degrading overall performance. Therefore, in this paper, we propose a \textcolor{black}{decentralized incentive mechanism for mobile AIGC service allocation}, employing multi-agent deep reinforcement learning to find the balance between the supply of AIGC services on RSUs and user demand for services within the IoV context, optimizing user experience and minimizing transmission latency. Experimental results demonstrate that our approach achieves superior performance compared to other baseline models.
\end{abstract}


\section{Introduction}
Artificial Intelligence-Generated Content (AIGC) is generally referred to as the use of AI models that automate the generation of content to match the requirements of users, which involves the creation of content such as video, audio, and text-based content\cite{cao2023comprehensive,du2023enabling}. By extracting and learning semantic information from human input, AIGC models could generate context-based content according to the intention of the user. Furthermore, they have the potential to offer high-quality content-generation services with much better efficiency and accessibility than their human counterparts, making them an attractive alternative in domains such as graphic design, customer service, and education. 

\textcolor{black}{Mobile AIGC services can greatly enhance the user experience on IoV infotainment systems, with which, users can enjoy entertainment services, monitor vehicle conditions, and perform real-time queries about traffic conditions or any matters of interest\cite{fan2022differentiated}. Moreover, the infusion of Mobile AIGC services enhances the infotainment system's content by skillfully generating a diverse array of responsive and engaging content for user queries. For example, AIGC-based entertainment services such as Changya (AI-aided music creation) and Faceplay (AI face-swap video app) \cite{thomala_2023} enrich the variety of in-car audiovisual services, where users could give the reins to their imagination instead of being restricted by limited media content from traditional in-car karaoke or video makers.}
%From a functional perspective, AIGC-based models could be trained using local traffic and driver behaviour data to provide real-time navigation plans and respond to traffic-related queries in a speedy and location-aware manner\cite{MASSAI201970}. In the event of a medical emergency, drivers could use such services to obtain the best route to the nearest hospital emergency department and use the AIGC-based dialer to make automatic calls with the emergency hotlines and exchange information about the present situation. This improves both traffic safety and communication efficiency, where drivers spend less time waiting and explaining the situation to emergency medical services over the phone to get directions on where to go, and medical teams can obtain all necessary information for immediate medical service upon their arrival.

However, AIGC services typically impose resource-intensive demands. As a result, delivering mobile AIGC services within latency-sensitive contexts like IoVs confronts substantial challenges related to both resource demand and latency requirements. To meet the requirements of mobile AIGC services, mobile AIGC networks have been envisioned as a feasible solution~\cite{xu2023unleashing}. In mobile AIGC networks, models are deployed at roadside units (RSUs), servicing incoming local user requests. 
% It has the advantages of better network efficiency by processing large volumes of data locally, higher reconfigurability by continuously adjusting models to meet local preferences, and improved data security and privacy by storing and processing sensitive user data locally rather than uploading them to remote locations~\cite{xu2023unleashing}. For AIGC service provisioning in mobile environments such as the IoV network, RSUs deployed along the traffic infrastructure could possibly virtualize their resources and host independent virtual machines to provide appropriate environments for different AIGC models. 
Nonetheless, the heterogeneous pool of AIGC models may involve different resources for training, storing, and fine-tuning. Given the cost of maintenance and service provision, RSUs would need a strategy to optimize their resource allocation amongst the pool of AIGC services that they provide. Furthermore, the limited computation and network resources available for fulfilling user requests in these RSUs lead to competition for services among users~\cite{9003106}. 
Hence, another significant challenge is determining how to appropriately allocate services among user requests to improve the overall user experience while maintaining low latency.

To address the aforementioned challenges, this paper aims to develop a decentralized service allocation mechanism by leveraging multi-agent deep reinforcement learning (MADRL)~\cite{madrl} to learn and balance the continuous supply and demand between AIGC services of RSUs and the user requests for services in decentralized IoV, optimizing the overall user experience and minimizing transmission latency.
In specific, each RSU and the IoV within the RSU's geographical proximity are modelled as a single local market. In each local market, each virtual machine maintains a different AIGC model and acts as a seller of the AIGC service the model provides, and each IoV acts as a potential buyer of the services. Thus, the goal of our mechanism is to match the supply of AIGC services and the demand for services from the user, optimizing overall user satisfaction, i.e., the accuracy of the model output in terms of summarizing quality\cite{Park2021BenchmarkFC} and service latency.

  \begin{figure}[htb]
    \begin{center}
        \includegraphics[width=1\linewidth]{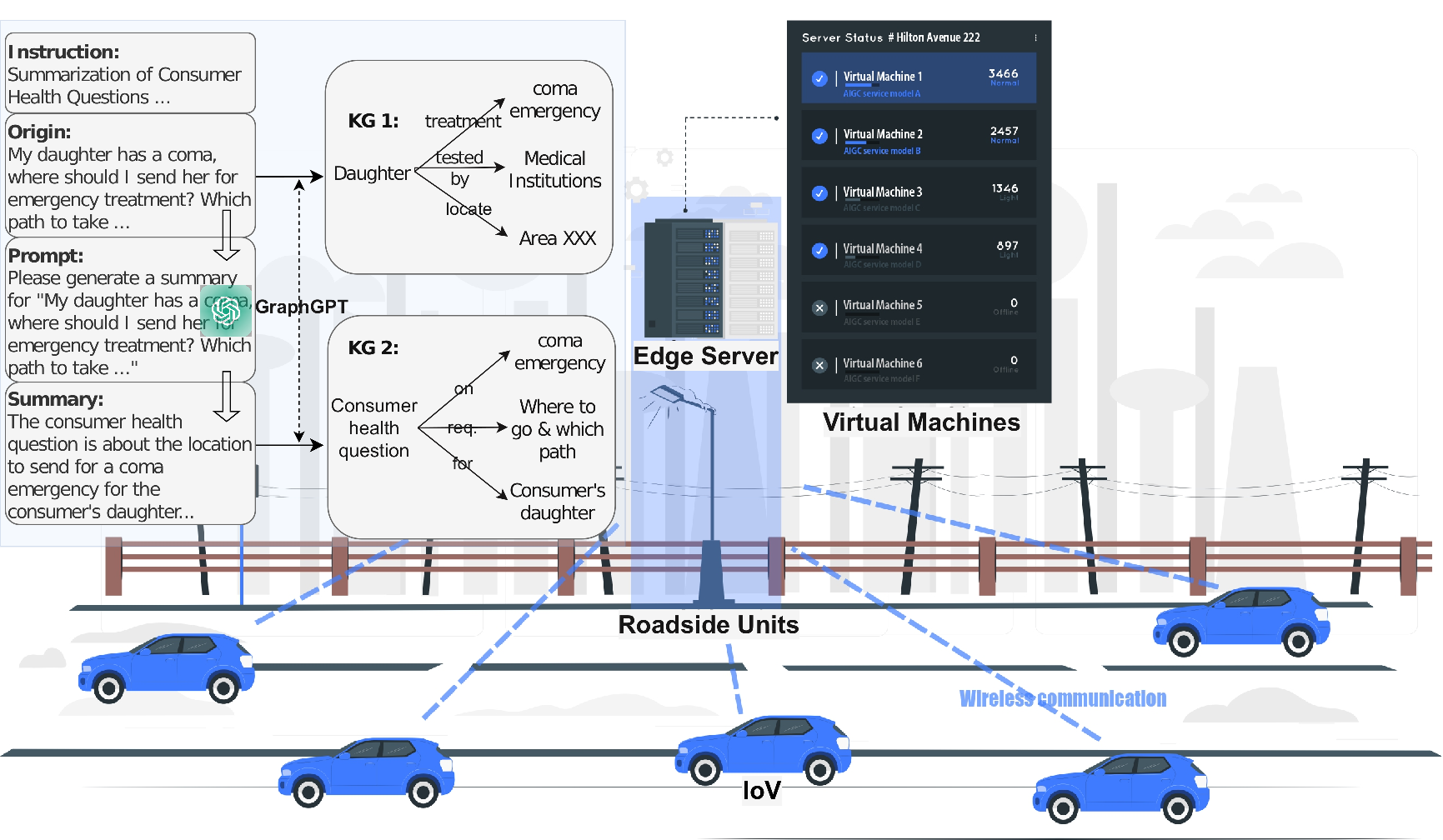}
        \caption{\textcolor{black}{IoVs competing for local AIGC services provided at the nearest RSU.}}
        \label{fig: overview}
    \end{center}
    \vspace{-5mm}
    \end{figure}
    
The main contributions of the paper are summarized as follows:

\begin{itemize}
\item We propose the market design of the continuous AIGC service allocation in decentralized IoV, where the AIGC service allocation in the local market is modeled as a double-sided model.
\item We propose a multi-agent deep reinforcement learning-based mechanism for continuous service allocation in the decentralized IoV network, optimizing overall user satisfaction while reducing total service latency.
\item We conduct experiments to evaluate the performance of our mechanism. The proposed mechanism can achieve satisfactory performance in comparison to the baseline models in the hierarchical double-side market.
\end{itemize}

% The remainder of the paper is organized as follows. We present an overview of our approach and problem definition in Section \ref{sec: overview}, details on the proposed multi-agent deep reinforcement learning-based service allocation mechanism in Section \ref{sec: rl}, experimental results of our mechanism in Section \ref{sec: experiments}, and a conclusion in Section \ref{sec: conclusion}.

% \section{Literature Review}
% \subsection{AIGC}

% \subsection{IoV}

\section{System Overview and Problem Definition} \label{sec: overview}
    
AIGC services often require large amounts of computing and storage resources for training and maintenance. Hence, many AIGC models are trained on remote cloud servers and exhibit high service latency. A feasible solution for reducing service latency is to deploy these models on RSUs for \textcolor{black}{servicing local user requests, where these models can also improve their location awareness with continuous input from local IoV network formed by nearby IoV users}.

An overview of the system is presented in Fig. \ref{fig: overview}, where we consider a decentralized continuous market with $T$ time slots in the system $T = \{1, \dots, T\}$. In the IoV network, RSUs are typically deployed to serve service requests from IoVs within their coverage. At each time slot, the IoV can choose to participate in the local market and compete for the AIGC service with other local IoVs or exit the local market to obtain maximum utility. Due to the limited resources at each RSU, the interaction is modelled as a non-cooperative game where IoVs compete to maximize their utility with minimum credit spending, i.e., achieving the highest user satisfaction in terms of service latency and output accuracy with the minimum price paid to the service sellers (i.e., virtual machines).

There are three main roles of participants in the system, i.e.,  RSUs, virtual machines on RSUs, and IoVs. More details about these participants can be found in the subsection below.

\subsection{Market Design}
In the context of the IoV network, we consider a decentralized global market where each RSU $n\in\mathcal{N}$ acts as an auctioneer for its local market to hold an auction between the set of sellers, i.e., the virtual machines $\mathcal{M}=\{1,\ldots, M\}$ that sell different AIGC services, and the set of buyers, i.e., the IoVs $\mathcal{V}=\{1,\ldots, V\}$ that request AIGC services. Hence, this market is double-sided, where bids from multiple buyers and sellers need to be aggregated before the service allocation and pricing~\cite{niyato2020auction}. Specifically, for each time slot $T$ in $T = \{1, \dots, T\}$, buyers and sellers in the local market submit their bids to their RSU, which decides on the service allocation and prices using the auction-based mechanism.

\subsubsection{\textbf{RSUs}}
RSUs possess considerable computing and GPU memory resources that can maintain extensive AIGC models and fine-tune them to align with local preferences. Each RSU $n\in\mathcal{N}$ fulfils the service requests of the nearby IoVs. Due to the variety of commercial AIGC services that users can choose from, RSUs must optimize their resource allocation to maximize user satisfaction with their limited resources.

\textcolor{black}{For instance, AIGC service providers can host $M$ AIGC models in each RSU, each in a different virtual environment such as a container or virtual machine.} However, each RSU has a limited amount of resources to provide $M$ AIGC services that have different demands for computation, network, and storage resources. Hence, the RSU may not be able to fulfil all incoming user requests without degrading the quality of AIGC services. \textcolor{black}{An effective strategy would be to model each RSU as a local market where virtual machines hosted by AIGC service providers are service sellers and IoVs are buyers in the market.} By finding the balance between the demand and supply of AIGC services through auction mechanisms, the RSU can find the optimal allocation of services to maximize the social welfare of the AIGC market.

\subsubsection{\textbf{Virtual Machines}}
In each RSU $n\in\mathcal{N}$, we consider a set of $M$ virtual machines where each virtual machine hosts an AIGC model $\{a^{i}$ in $a^{1},\ldots, a^{m}\}$. These $M$ virtual machines act as independent sellers of AIGC services, where they can fulfil a limited number of service requests without affecting the quality of service. In the system, each virtual machine $m\in \mathcal{M}$ on the RSU aims to earn a profit from the IoV by satisfying the user request with the required service quality, e.g., model accuracy. At the same time, the virtual machine $m$ has a valuation of $u_m(t)$ for the AIGC service it can provide at time slot $t$. This valuation is affected by the transmit power $p_m(t)$ and the cost $c^{m}(t)$ for providing the service using model $a^{m}(t)$, where $c^{m}(t)$ includes the average computation cost $c_{c}^{m}(t)$, the size of the content of the service $c_{s}^{m}(t)$, and the storage cost $c_{r}^{m}(t)$ to provide one unit of service using the model $a^{m}(t)$. In addition, the valuation follows the economic law of ``increasing marginal costs” \textcolor{black}{where the cost of producing additional units of service increases with more production}. Hence, the utility function of the virtual machine $m\in\mathcal{M}(t)$ at time slot $t$ can be represented as
$ \mu_m(t)(b_m(t)) = p_m(t) - u_m(t), $
where $b_m(t)$ is the selling bid of virtual machine $m$ and $p_m(t)$ is the selling revenue of virtual machine $m$ at time slot $t$.

\subsubsection{\textbf{IoVs}}
In the local market $n$, there are $V$ IoVs travelling at different speeds under the coverage of different RSUs. At each time slot $t$, the IoVs can choose to request an AIGC service from the nearest RSU, i.e., participate in the local market. Each IoV that requests service from the virtual machines wants to satisfy its request as fast and accurately as possible, at the lowest possible price, or both. At a given time slot $t$, if the IoV $v$ participates in the local market $n$ then $v$ determines its valuation $u_v(t)$ for the service and $g_{v,n}(t)=1$, and otherwise $g_{v,n}(t)=0$. This valuation is affected by service latency and accuracy, e.g., Frechet Inception Distance (FID)\cite{Park2021BenchmarkFC} and the quality of generated texts, and follows the economic law of ``diminishing marginal returns". Therefore, the utility function of requesting IoV $v\in\mathcal{V}(t)$ can be represented as $\mu_v(t)(b_v(t)) = u_v(t) - p_v(t),$ where $b_v(t)$, $p_v(t)$ are the buying bid and charge of IoV $v$ respectively.

After submitting the purchase bid to the central auctioneer, the winning IoVs $v \in \mathcal{V}$ receive the AIGC service they requested from the virtual machines $m\in \mathcal{M}$ with a transmission rate of $R_{m,v}^{d}(t)$, and the losing IoVs can choose to bid in the next time slot $t+1$ or later if they decide to participate again.

\subsection{Problem Definition}
To determine the global market equilibrium for the supply and demand of AIGC services, the RSU needs to establish allocation and pricing rules that satisfy the constraints of individual rationality (IR) and truthfulness. Individual rationality means that the sellers and buyers receive non-negative returns from their participation in the local market, i.e., IoVs are paying a price less than their valuation, i.e., $p_v(t) \leq u_v(t)$ and virtual machines are receiving a payment greater than its valuation, i.e., $p_m(t) \geq u_m(t)$. Truthfulness refers to the situation of whether sellers' and buyers' bid prices reflect their true valuation of the services. For example, a truthful bid of buyer $v$ $b_v(t)$ corresponds to their true valuation $u_v$, indicating that $b_v(t) = u_v(t)$. A truthful auction mechanism must ensure that payment $p_v(t)$ of buyer $v$ is not related to its true valuation $u_v$. For sellers, a truthful ask $a_m(t)$ represents their true valuation of $u_m(t)$, such that $a_m(t) = u_m(t)$. The auction mechanism needs to guarantee that its revenue $p_m(t)$ is not calculated based on its bids. Then, the pricing rules can be represented by the pricing vector of buyers $\mathbf{p}^v(t)$ and the pricing vector of sellers $\mathbf{p}^m(t)$.

Then, we explain how each RSU can calculate its local allocation and pricing rules for the auction in time slot $t$. In specific, each RSU can calculate its local allocation and prices in the time slot $t$ based on the auction mechanism $\mathcal{Z}=(\Pi, \Psi)$, where $\Pi=(X(t), Y(t))$ is the allocation rules and $\Psi=(\mathbf{p}^v(t), \mathbf{p}^m(t))$ is the pricing rules. The allocation rule $\Pi$ consists of the supply matrix $X(t)$ of the virtual machine $m$ where $X(t) \subseteq \{0,1\}^{|\mathcal{M}(t)|} = \{\mathbf{x}_1(t),\ldots,\mathbf{x}_{|\mathcal{M}(t)|}(t)\}$ with supply vector $\mathbf{x}_{m}(t) =\{x_{m,1}(t), \ldots, x_{m,|\mathcal{M}(t)|}(t)\}$,
and demand matrix $Y(t)$ of the IoV $v$ where $Y(t)\subseteq \{0,1\}^{|\mathcal{V}(t)|} = \{\mathbf{y}_1(t),\ldots,\mathbf{y}_{|\mathcal{V}(t)|}(t)\}$ with demand vector $\mathbf{y}_{v}(t) =\{y_{v,1}(t), \ldots, y_{v,|\mathcal{V}(t)|}(t)\}$.

To assess the effectiveness of the auction mechanism, we can use the global social welfare $SW$ as the main metric which represents the total utility gained by both sellers and buyers who have been assigned a service transaction. The maximum value of $SW$ is achieved when there is a balance between the supply and demand of AIGC services. This means that all demands for AIGC have been matched and satisfied with available service supplies from virtual machines. The global social welfare $SW(t)$ at time slot $t$ can be represented as
\begin{equation}
\begin{aligned}
    SW(t) =& \sum_{v\in \mathcal{V}(t)}\sum_{m\in \mathcal{M}(t)} x_{v,m}(t) u_v(t) \\&+ \sum_{v\in \mathcal{V}(t)} \sum_{m\in \mathcal{M}(t)} y_{v,m}(t) u_{m}(t).
\end{aligned}
\end{equation}

The total transmission latency $L(t)$ of AIGC services at time slot $t$ between virtual machines and IoVs can be calculated as
\begin{equation}
\begin{aligned}
    L(t) = &  \sum_{n \in \mathcal{N}} \sum_{v \in \mathcal{V}(t)} g_{n,v}(t) \sum_{m \in \mathcal{M}(t)}  x_{m,v}(t)\frac{c_{s}^{m}(t)}{R_{m,v}^{d}(t)}.
\end{aligned}
\end{equation}

To maximize global social welfare and minimize total service latency, the optimization problem can be defined as
\begin{maxi!}|s|[2]<b>
    {\mathcal{M}}{\frac{1}{T} \sum_{t\in \mathcal{T}} \big(SW(t) - L(t)\big)}{}{}
    \addConstraint{\sum_{n\in\mathcal{N}} g_{v,n}(t)}{=1,}{\forall v\in \mathcal{V}, t \in \mathcal{T}\label{con1}}{}{}
    \addConstraint{\sum_{v\in \mathcal{V}(t)} y_{m,v}(t)}{\leq 1,}{\forall m\in \mathcal{M}(t), t \in \mathcal{T} \label{con2}}{}{}
    \addConstraint{ x_{m,v}(t)}{\leq y_{m,v}(t),}{\forall v \in \mathcal{V}(t), m \in \mathcal{M}(t), t \in \mathcal{T} \label{con3}}{}{}.
\end{maxi!}

The constraint \eqref{con1} indicates that any IoV can participate in only one of the local markets at any time slot $t \in \mathcal{T}$. The constraint \eqref{con2} ensures that any service transaction at any time slot $t \in \mathcal{T}$ can only be performed once, and the constraint \eqref{con3} represents the allocation variables of the mechanism.

\section{MADRL-based Service Allocation Mechanism} \label{sec: rl}
In this mechanism, we consider each RSU $n\in\mathcal{N}$ as the auctioneer of its local market, where virtual machines on each RSU act as service sellers of the AIGC services that the service providers are hosting in these virtual machines, and the IoVs under the coverage of the RSU act as service buyers. At each time slot $t$, sellers $m\in\mathcal{M}(t)$ and buyers $v\in\mathcal{V}(t)$ who wish to participate in the service transaction submit their selling bids $d_{m}(t)$ and buying bids $d_{v}(t)$ to the auctioneer and wait for market clearance at the end of $t$, where buying and selling bids are matched. 

\subsection{Mechanism Design}
In each local market $n$, the sellers are organized into a seller pool $\mathcal{P}^M_{n}(t) = \{ m | \sum_{m\in \mathcal{M}(t)} y_{m,v}(t) = 0, m\in \mathcal{M}(t)\}$. The selling bids of the sellers are organized into a seller's value pool $\mathcal{D}^M_{n}(t) = \{ d_{m}(t) | \sum_{v\in \mathcal{V}(t)} y_{m,v}(t) = 0, m\in \mathcal{M}(t)\}$, which is sorted by their bids $d_{m}(t) \leq d_{m+1}(t), \forall m, m+1 \in \mathcal{P}^M_{n}(t)$. The buyers are organized into a buyer pool $\mathcal{P}^V_{n}(t) = \{v|g_{v,n}=1, \sum_{m\in \mathcal{M}(t)} x_{v,m}(t)=0, v\in \mathcal{V}(t)\}$. The buying bids of the buyers are organized into a buyer value pool $\mathcal{D}^V_{n}(t)$, which is sorted by their bids $d_{v}(t) \geq d_{v+1}(t),  \forall v, v+1 \in \mathcal{P}^V_{n}(t)$. Then, the auctioneer determines the allocation and the pricing rule leveraging McAfee's mechanism~\cite{mcafee1992dominant}. 

At time slot $t$, the auctioneer finds the breakeven index $K$ in $\mathcal{D}^M_{n}(t)$ and $\mathcal{D}^V_{n}(t)$ for the set of seller $\mathcal{P}^M_{n}(t)$ and buyer $\mathcal{P}^V_{n}(t)$ pool. Then, the auctioneer calculates the average price $p(t)=(v_{k+1}+m_{k+1})/2$. If $\sum_{v\in \mathcal{P}^V_{n}(t)}1_{\{b_v(t) \geq p(t)\}} = K$ and $\sum_{m\in \mathcal{P}^M_{n}(t)}1_{\{a_m(t) \leq p(t)\}} = K$, then the allocation rules for the first $K$-th buyers and sellers are set to $x_{v_k, m_k}(t) = 1$ with a clearing price of $p_{v_k}(t) = p_{m_k}(t) = p(t)$.
Otherwise, the first $K-1$-th sellers trade for $m_k$ and the first $k-1$ buyers trade for $v_k$ as in the trade-reduction mechanism with the allocation rules of $x_{v_k, m_k}(t) = x_{v_k, m_k}(t) = 1$ and a clearing price of $p_{v_k}(t) = p_{K}^v(t)$ and $p_{m_k}(t) = p_{K}^m(t)$ respectively.

After completing the clearance of the local markets, the auctioneer $n$ measures its local budget cost $\beta_n(t)$, which can be calculated as 
\begin{equation}
\begin{aligned}
    \beta_n(t) = &
    \sum_{v\in\mathcal{V}(t)} g_{v,n}(t) x_{v,m}(t) p_v(t) - \sum_{m\in\mathcal{M}(t)} y_{v,m}(t) p_m(t).
\end{aligned}
\end{equation}
Following that, the total budget cost of the global decentralized market is the sum of the local budget cost $\beta_n(t), \forall n\in\mathcal{N}$, which can be calculated as $\beta(t) = \sum_{n\in\mathcal{N}} \beta_n(t)$.

To minimize the total budget cost and maximize social welfare, we trained multiple learning agents to represent buyers in learning the optimum bidding strategy in the decentralized market to find market equilibrium.

\subsection{Multi-agent Reinforcement Learning-based Mechanism}
Under the proposed mechanism design, each buyer $v\in\mathcal{V}$ is represented with a reinforcement learning agent in a Partially observable Markov decision process (POMDP) that has the following components:

\subsubsection{Observation} The observation $O_v(t)$ of IoV $v$ at time step $t$, which represents the state $S(t)$, includes the number of buyers and sellers in each local market $|V_n(t)\bigcup M_n(t)|, \forall n \in \mathcal{N}$, the last transaction price $\bar{p}(t-1)$, and the transmission rate between buyers and sellers $R_{m,v}^{d}(t)$. This observation can be formulated as
\begin{equation}
\begin{aligned}
    O_v(t) = \{|V_1(t)\cup M_1(t)|, &\ldots, |V_N(t)\cup M_N(t)|, \\&\bar{p}(t-1), R_{m,v}^{d}(t)\}
\end{aligned}
\end{equation}
\subsubsection{Action} The action of IoV $v$ at time slot $t$ refers to the determination of buying bid $b_v(t)$ the IoV submits to the auctioneer.
\subsubsection{Reward} The reward function includes global social welfare, total budget costs, and transmission latency at the current time slot $t$, which is represented as
\begin{equation}
R_v(S(t), b_v(t)) = SW(t) - \alpha \beta(t)^2 - L(t),
\end{equation}
where $\alpha$ is the coefficient of the total budget cost.
Specifically, global social welfare $SW(t)$ is the total utility of all buyers and sellers in the system as a function of the transaction prices. The total budget cost $\beta(t)$ is the difference between the total payment received from buyers and the total price paid to sellers from the transactions. The transmission latency $L(t)$ is the total delay incurred due to the transmission of service content between sellers and buyers.

\subsubsection{Valuation Function} Given the policy $\pi_v$ of IoV $v$ and the valuation function $V_{\pi_v}(S)$ of state $S(t)$, the expected return with a global state $S$ can be represented as
	\begin{equation}
	V_{\pi_v}(S) := \mathbb{E}_\pi\left[\sum_{t=0}^{T}\gamma^k R_v(S(t), b_v(t))|S^0=S\right],
	\end{equation}
	where $\mathbb{E}_\pi(\cdot)$ denotes the expected value of a random variable given that the learning agent follows policy $\pi$. $\gamma\in[0,1]$ is the discount factor on rewards for reducing the weights as the time step increases.

The POMDP framework provides a mathematical model for optimizing the buyer's bidding decisions, where buyers observe the state of the systems and make appropriate actions to maximize the expected reward. In addition, the POMDP framework can be extended for learning agents to maximize individual utility while improving global social welfare under the Multi-agent Proximal Policy Optimization (MAPPO) algorithm, which trains multiple agents simultaneously. 

In this framework, IoV $v$ decides the best action under its policy $\pi_{v}(\theta_v)$ to maximize its rewards according to the current observation of the system state. Let $\theta_v$ be the policy parameter of IoV $v$ and $\theta_v$ be the parameter for the valuation function. With MAPPO, each agent is trained using the POMDP framework and maintains its policy $\pi_{v}(\theta_v)$, which is updated using a critic function $V(s;\phi_v)$ that estimates the valuation of the global state $S$. In addition, the critic $V(s;\phi_v)$ and policy $\pi_{v}(\theta_v)$ network is trained using the clip loss $L^\emph{CLIP} (\theta_v, \phi_v)= L^\emph{P}(\theta_v) +  L^\emph{V}(\phi_v)$, which consists of the loss of policy networks and valuation networks~\cite{schulman2017proximal} in the global state. \textcolor{black}{Furthermore, the proposed learning-based mechanism is IR and truthful in the decentralized market following the properties of the second-price auction and Macfee's double auction~\cite{niyato2020auction}.}

\begin{figure*}[!htb]
\minipage{0.24\textwidth}
  \includegraphics[width=\linewidth]{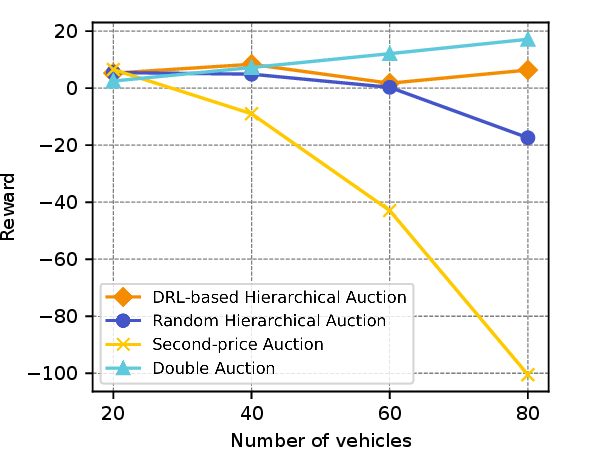}
  \caption{Reward {\color{black}versus} number of IoVs.} \label{fig:reward}
\endminipage\hfill
\minipage{0.24\textwidth}
  \includegraphics[width=\linewidth]{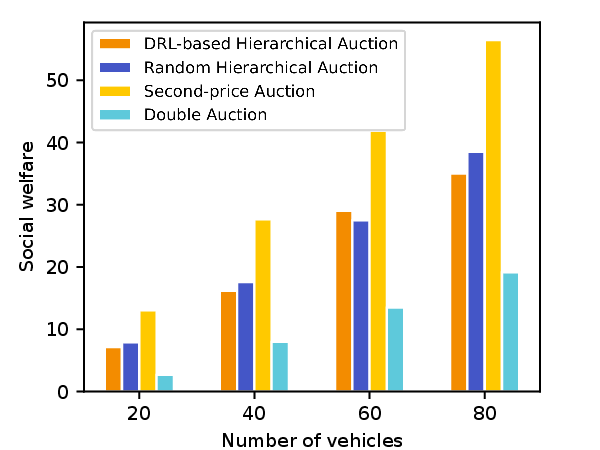}
  \caption{Social welfare {\color{black}versus} number of IoVs.} \label{fig:social}
\endminipage\hfill
\minipage{0.24\textwidth}%
  \includegraphics[width=\linewidth]{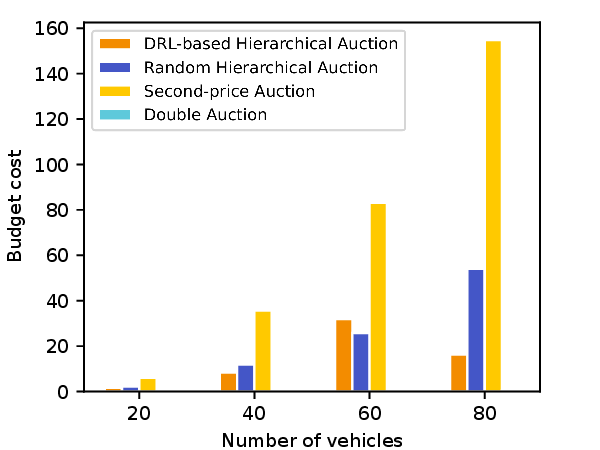}
  \caption{Budget cost {\color{black}versus} number of IoVs.} \label{fig:budget}
\endminipage\hfill
\minipage{0.24\textwidth}%
  \includegraphics[width=\linewidth]{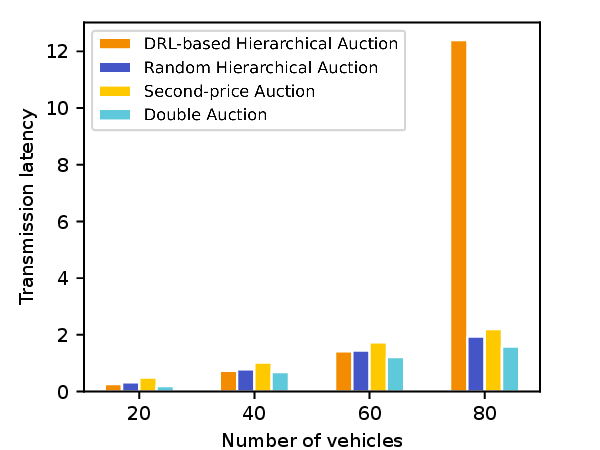}
  \caption{Transmission latency (ms) {\color{black}versus} number of IoVs.} \label{fig:latency}
\endminipage
\vspace{-5mm}
\end{figure*}

\section{Evaluation Results} \label{sec: experiments}
To simulate the IoV environment, a 1km$\times$1km transportation network with 4 RSUs is considered, each covering a range of 500 m~\cite{liang2019spectrum}. IoVs are travelling on the network at an average speed of 90 km per hour, and their roles on the market are determined based on their requests at current time slots. The amount of AIGC service requested by IoV $v$ is evaluated from the ChatGPT simulation in the MeQSum dataset~\cite{MeQSum} and its valuation $u_v$ is formulated as $\log(1+c_{s}^{m}/1000)$. On the other hand, the virtual machine $m$ \textcolor{black}{on RSU $n$} has a transmission power of $p_{v}^{m}$ sampled from $U[0,10]$ and a utility value $u_m = p_{v}^{m}/10$ proportional to its transmission power. We evaluated the performance of our approach in an environment simulated with $20, 40, 60,$ and $80$ IoVs. The learning rate of our DRL agents is set to $0.001$ with a discount factor of $0.95$. The coefficients of the entropy function and the value function are set to $0.02$ and $0.5$ respectively, with a clip factor of $0.2$.

% Initial experiments are conducted to ascertain that the MADRL-oriented approach attains a commendable level of performance within the decentralized, continuous data-sharing marketplace. 
% As shown in Fig. \ref{fig:convergence}, 
With experiments, our algorithm can outperform the double auction-based mechanism in 10 epochs of training and reaches an acceptable reward threshold after roughly 25 epochs, demonstrating both efficiency and effectiveness. In addition, the market rewards have improved by up to 10\% due to the MADRL algorithm's capacity to learn from historical data and prevailing market conditions. Figure \ref{fig:reward} demonstrated the rewards obtained from our method and alternative baselines. With fewer IoVs, the second-price auction has the best rewards, but its efficacy declines steeply as the number of IoVs increases. The random approach has a comparable trajectory to the second-price auction, but it cannot fully utilize the contextual information in the market, \textcolor{black}{such as historical prices and current market conditions}. \textcolor{black}{On the other hand, our double auction mechanism consistently delivers satisfactory and stable performance under various market conditions.}

In terms of social welfare, shown in Fig. \ref{fig:social}, the second-price auction outperforms all other mechanisms due to its efficient resource allocation from honest bidding. In contrast, the social welfare gained from the double auction fluctuates depending on the number of IoVs due to its increasing complexity and intensified bidder interaction. Meanwhile, our mechanism and random mechanism have comparable social welfare gains.

In Fig. \ref{fig:budget}, we can observe that the number of IoVs significantly affects the budget cost for the second-price auction due to the involvement of an increasing number of operations, thereby generating increased communication and computation costs. In contrast, the budget cost of the double auction is always zero, as it has no communication nor computation needs. The random mechanism has a lower budget cost than the DRL-based mechanism due to its simplicity, which forgoes any computational or communicative requirements.

In Fig.~\ref{fig:latency}, a comparison of transmission latency for the proposed and baseline mechanisms is presented. As we can observe, the double auction has the lowest transmission delay as it accommodates a smaller volume of transactions, regardless of the number of IoVs. Nonetheless, it is crucial to recognize that the double auction may not have the highest social welfare despite achieving the lowest budget cost. In contrast to the random mechanism, the proposed DRL-based mechanism reduces transmission latency by harnessing the capabilities of deep reinforcement learning to develop an optimal bidding function that minimizes transmission latency while achieving a satisfactory level of social welfare.

% \begin{figure}[t]
%     \centering
%     \includegraphics[width=0.8\linewidth]{AIGC_reward.pdf}
%     \caption{Reward {\color{black}versus} number of IoVs.}
%     \label{fig:reward}
%     \vspace{-5mm}
% \end{figure}

% \begin{figure}[t]
%     \centering
%     \includegraphics[width=0.8\linewidth]{AIGC_socialwelfare.pdf}
%     \caption{Social welfare {\color{black}versus} number of IoVs.}
%     \label{fig:social}
%     \vspace{-5mm}
% \end{figure}

% \begin{figure}[t]
%     \centering
%     \includegraphics[width=0.8\linewidth]{AIGC_budget.pdf}
%     \caption{Budget cost {\color{black}versus} number of IoVs.}
%     \label{fig:budget}
%     \vspace{-5mm}
% \end{figure}

% \begin{figure}[t]
%     \centering
%     \includegraphics[width=0.8\linewidth]{AIGC_latency.pdf}
%     \caption{Transmission latency {\color{black}versus} number of IoVs.}
%     \label{fig:latency}
%     \vspace{-5mm}
% \end{figure}

\section{Conclusion} \label{sec: conclusion}

In this paper, we have proposed a decentralized mechanism for AIGC service allocation in the IoV context. By employing multi-agent reinforcement learning, our approach can find the market balance in the supply and demand of AIGC services by the virtual machine and IoVs, optimizing user satisfaction and minimizing transmission latency. With experimentation, our approach can achieve a good tradeoff in terms of budget cost, total social welfare, and transmission latency.

\bibliographystyle{bibstyle}

\bibliography{references}

\begin{thebibliography}{10}

\bibitem{cao2023comprehensive}
Y.~Cao, S.~Li, Y.~Liu, Z.~Yan, Y.~Dai, P.~S. Yu, and L.~Sun, A Comprehensive
  Survey of AI-Generated Content (AIGC): A History of Generative AI from GAN to
  ChatGPT 2023.

\bibitem{du2023enabling}
H.~Du, Z.~Li, D.~Niyato, J.~Kang, Z.~Xiong, Xuemin, Shen, and D.~I. Kim,
  Enabling AI-Generated Content (AIGC) Services in Wireless Edge Networks 2023.

\bibitem{fan2022differentiated}
J.~Fan, L.~K. Shar, J.~Guo, W.~Yang, D.~Niyato, and K.-Y. Lam, Differentiated
  security architecture for secure and efficient infotainment data
  communication in IoV networks in {\em International Conference on Network and
  System Security}, pp.~283--304, Springer, 2022.

\bibitem{thomala_2023}
L.~L. Thomala, China: Popular AIGC apps 2022 Apr 2023.

\bibitem{xu2023unleashing}
M.~Xu, H.~Du, D.~Niyato, J.~Kang, Z.~Xiong, S.~Mao, Z.~Han, A.~Jamalipour,
  D.~I. Kim, Xuemin, Shen, V.~C.~M. Leung, and H.~V. Poor, Unleashing the Power
  of Edge-Cloud Generative AI in Mobile Networks: A Survey of AIGC Services
  2023.

\bibitem{9003106}
Z.~Cheng, Q.~Wang, Z.~Li, and G.~Rudolph, Computation Offloading and Resource
  Allocation for Mobile Edge Computing in {\em 2019 IEEE Symposium Series on
  Computational Intelligence (SSCI)}, pp.~2735--2740, 2019.

\bibitem{madrl}
W.~Du and S.~Ding, A survey on multi-agent deep reinforcement learning: from
  the perspective of challenges and applications {\em Artificial Intelligence
  Review}, vol.~54, pp.~1--24, 06 2021.

\bibitem{Park2021BenchmarkFC}
D.~H. Park, S.~Azadi, X.~Liu, T.~Darrell, and A.~Rohrbach, Benchmark for
  Compositional Text-to-Image Synthesis in {\em NeurIPS Datasets and
  Benchmarks}, 2021.

\bibitem{niyato2020auction}
D.~Niyato, N.~C. Luong, P.~Wang, and Z.~Han, Auction theory for computer
  networks 2020.

\bibitem{mcafee1992dominant}
R.~P. McAfee, A dominant strategy double auction {\em Journal of economic
  Theory}, vol.~56, no.~2, pp.~434--450, 1992.

\bibitem{schulman2017proximal}
J.~Schulman, F.~Wolski, P.~Dhariwal, A.~Radford, and O.~Klimov, Proximal policy
  optimization algorithms {\em arXiv preprint arXiv:1707.06347}, 2017.

\bibitem{liang2019spectrum}
L.~Liang, H.~Ye, and G.~Y. Li, Spectrum sharing in vehicular networks based on
  multi-agent reinforcement learning {\em IEEE Journal on Selected Areas in
  Communications}, vol.~37, no.~10, pp.~2282--2292, 2019.

\bibitem{MeQSum}
A.~{Ben Abacha} and D.~Demner-Fushman, On the Summarization of Consumer Health
  Questions in {\em Proceedings of the 57th Annual Meeting of the Association
  for Computational Linguistics, ACL 2019, Florence, Italy, July 28th - August
  2}, 2019.

\end{thebibliography}

\end{document}